\documentclass[sigconf]{acmart}
\usepackage{booktabs}
\usepackage{multirow}
\usepackage{subfigure}
\usepackage{bm}
\usepackage{cleveref}
\usepackage{hyperref}
\hypersetup{colorlinks=true, citecolor=green}

\AtBeginDocument{%
  \providecommand\BibTeX{{%
    \normalfont B\kern-0.5em{\scshape i\kern-0.25em b}\kern-0.8em\TeX}}}

\renewcommand\footnotetextcopyrightpermission[1]{}
\acmSubmissionID{1347}

\begin{document}

\title{Food-500 Cap: A Fine-Grained Food Caption Benchmark for Evaluating Vision-Language Models}

\author{Zheng Ma}
\affiliation{%
  \institution{National Key Laboratory for Novel Software Technology,\\ Nanjing University,}
  \city{Nanjing}
  \country{China}}
\email{maz@smail.nju.edu.cn}
\authornote{Equal contribution.}

\author{Mianzhi Pan}
\affiliation{%
  \institution{National Key Laboratory for Novel Software Technology,\\ Nanjing University,}
  \city{Nanjing}
  \country{China}}
\email{panmz@smail.nju.edu.cn}
\authornotemark[1]

\author{Wenhan Wu}
\affiliation{%
  \institution{National Key Laboratory for Novel Software Technology,\\ Nanjing University,}
  \city{Nanjing}
  \country{China}}
\email{wuwh@smail.nju.edu.cn}

\author{Kanzhi Cheng}
\affiliation{%
  \institution{National Key Laboratory for Novel Software Technology,\\ Nanjing University,}
  \city{Nanjing}
  \country{China}}
\email{chengkz@smail.nju.edu.cn}

\author{Jianbing Zhang}
\affiliation{%
  \institution{National Key Laboratory for Novel Software Technology,\\ Nanjing University,}
  \city{Nanjing}
  \country{China}}
\email{zjb@nju.edu.cn}
\authornote{Corresponding author.}

\author{Shujian Huang}
\affiliation{%
  \institution{National Key Laboratory for Novel Software Technology,\\ Nanjing University,}
  \city{Nanjing}
  \country{China}}
\email{huangsj@nju.edu.cn}
\authornotemark[2]

\author{Jiajun Chen}
\affiliation{%
  \institution{National Key Laboratory for Novel Software Technology,\\ Nanjing University,}
  \city{Nanjing}
  \country{China}}
\email{chenjj@nju.edu.cn}

\renewcommand{\shortauthors}{Zheng Ma et al.}

\begin{abstract}
Vision-language models (VLMs) have shown impressive performance in substantial downstream multi-modal tasks.
However, only comparing the fine-tuned performance on downstream tasks leads to the poor interpretability of VLMs, which is adverse to their future improvement.  
Several prior works have identified this issue and used various probing methods under a zero-shot setting to detect VLMs’ limitations, but they all examine VLMs using general datasets instead of specialized ones. 
In practical applications, VLMs are usually applied to specific scenarios, such as e-commerce and news fields, so the generalization of VLMs in specific domains should be given more attention. 
In this paper, we comprehensively investigate the capabilities of popular VLMs in a specific field, the food domain. 
To this end, we build a food caption dataset, Food-500 Cap, which contains 24,700 food images with 494 categories. Each image is accompanied by a detailed caption, including fine-grained attributes of food, such as the ingredient, shape, and color. We also provide a culinary culture taxonomy that classifies each food category based on its geographic origin in order to better analyze the performance differences of VLM in different regions.
Experiments on our proposed datasets demonstrate that popular VLMs underperform in the food domain compared with their performance in the general domain. Furthermore, our research reveals severe bias in VLMs' ability to handle food items from different geographic regions. We adopt diverse probing methods and evaluate nine VLMs belonging to different architectures to verify the aforementioned observations. 
We hope that our study will bring researchers' attention to VLM's limitations when applying them to the domain of food or culinary cultures, and spur further investigations to address this issue.
\end{abstract}





\maketitle

\section{Introduction}
Despite the remarkable success of vision-language models (VLMs)~\cite{DBLP:conf/iclr/SuZCLLWD20,DBLP:conf/emnlp/TanB19,DBLP:conf/icml/RadfordKHRGASAM21,DBLP:conf/icml/0001LXH22,DBLP:conf/icml/WangYMLBLMZZY22,DBLP:conf/icml/RameshPGGVRCS21,DBLP:conf/cvpr/RombachBLEO22} in substantial uni-modal and multi-modal downstream tasks, they are still poorly understood as yet. 
The prevalent approach for evaluating VLMs is comparing their performance on downstream tasks after fine-tuning.
However, evaluation solely based on the fine-tuning results renders poor interpretability \cite{DBLP:journals/corr/abs-2207-00221}, which hinders the further development of VLMs.
Consequently, researchers have proposed a range of probing methods and benchmarks ~\cite{DBLP:conf/acl/HendricksN21,DBLP:conf/naacl/RoschL22,DBLP:conf/emnlp/JiKRSVHA22,DBLP:conf/cvpr/ThrushJBSWKR22,DBLP:conf/acl/ParcalabescuCMF22,DBLP:conf/nips/ParkALDR21} in recent years to assess the capabilities of VLMs from various perspectives, providing a more comprehensive understanding of these models.
However, these methodologies are still limited in the general domain. 
They typically construct evaluation benchmarks by employing images from widely used general-domain datasets and subsequently assigning hand-crafted textual annotations to these images. 
If VLMs perform well in a specific domain, we can directly employ the models in that domain without any modifications. However, the above situation is unclear due to only few works studying the generalization of using VLMs directly in specific domains without fine-tuning.

\begin{figure}[t]
	\centering
	\includegraphics[width=.49\textwidth]{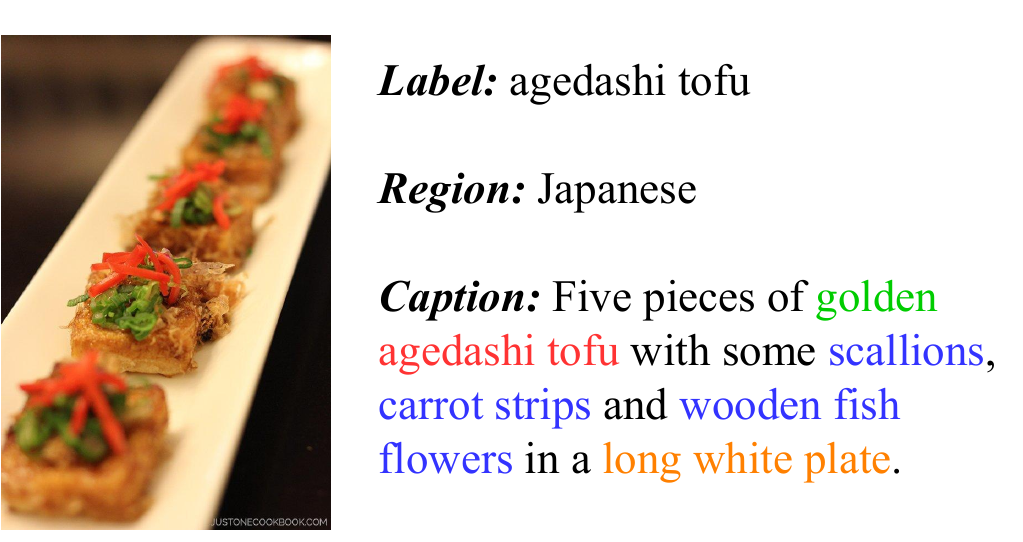}
	\caption{An example from our Food-500 Cap. The image is equipped with the label, geographic origin, and a detailed description. This description is annotated with a class label (red) and hand-curated various fine-grained visible content of the image such as ingredients (blue), food colors (green), and the food container (orange).}
	\label{fig:dataset_example}
\end{figure}
Motivated by this, we focus on evaluating the generalization capacity of VLMs in a specific domain, namely, the food domain. 
Since food computing~\cite{DBLP:journals/csur/MinJLRJ19} has been gaining widespread attention as it has the potential to support numerous food-related applications, such as healthy
diets and food choices. 
To comprehensively evaluate the VLMs' performance on food-related tasks, we introduce a new benchmark named Food-500 Cap, which comprises 24,700 food images with 494 categories, each accompanied by a detailed caption.
The Food-500 Cap dataset is created by selecting images from ISIA Food-500~\cite{DBLP:conf/mm/MinLWLWWJ20} that covers a wide range of food categories.
We select 50 images from each category and engage an annotation company to annotate fine-grained descriptions for all 24,700 images. 
Each description includes the original food category label and fine-grained attributes of the food, such as the color, shape, and ingredients. 
Such an in-house labeling process guarantees the high quality of our dataset. 
Besides, as food is always associated with a specific geographic region, we also provide a taxonomy that classifies food categories based on their original place, enabling a more comprehensive investigation of VLMs' performance across culinary cultures. We provide a sample of Food-500 Cap in \Cref{fig:dataset_example}, which contains a Japanese food image labeled \textit{agedashi tofu} from and a description with some related attributes.
In contrast to the prevalent food datasets~\cite{DBLP:journals/pami/MarinBOHSAWT21,DBLP:conf/mm/WuFLLHS21,DBLP:journals/corr/ChenZD17}, Food-500 Cap are equipped with  high-quality image captions containing richer visual information and geographic origin tags, which is more suitable for exploring the performance of VLM in the food domain.

To comprehensively evaluate VLMs' capacity in the food domain, we seriously pick up nine representative models from three popular architectures, including vision-language representation models (e.g. CLIP~\cite{DBLP:conf/icml/RadfordKHRGASAM21}), image-to-text generative models (e.g. OFA~\cite{DBLP:conf/icml/WangYMLBLMZZY22}), and text-to-image generative models (e.g. Stable Diffusion~\cite{DBLP:conf/cvpr/RombachBLEO22}). 
We probe these VLMs with various food-related tasks in a zero-shot setting. 
For vision-language representation models, we employ food classification and image-text retrieval to assess VLMs' multi-modal alignment capabilities. 
As for image-to-text generative models and text-to-image generative models, we utilize image captioning and image synthesis respectively to test their multi-modal generation capabilities. 
Both qualitative and quantitative analyses are performed on the experimental results, revealing that these models exhibit poor performance in the food domain, in contrast to their performance in the general domain.
Moreover, we find that all the models display a significant bias in culinary culture, with their performance in Asian cuisine falling markedly behind that in European, North American, and Latin American cuisine.
In summary, this paper makes the following contributions:

\begin{itemize}
    \item We equip a subset of the ISIA Food-500 dataset with (1) find-grained image descriptions (2) the geographic origin of each food category. Based on this enhanced dataset, we propose Food-500 Cap, which serves as a benchmark to evaluate the vision-language ability of VLMs in the food domain. To the best of our knowledge, Food-500 Cap is the first image-caption dataset that specifically targets the food domain.
    \item We evaluate nine representative VLMs from diverse architectures on our benchmark and use four probing tasks to analyze the performance of VLMs in the food domain comprehensively. 
    \item The results of our experiments on Food-500 Cap unveil the limitations of VLMs in the food domain, as well as their bias towards specific culinary cultures. 
\end{itemize}

\section{Related works}
\label{sec:related_works}
\subsection{Probing VLMs}
VLMs have achieved state-of-the-art performance in a large number of downstream multi-modal tasks, but they are still poorly understood. Therefore, evaluating VLMs has attracted much attention.
Commonly, VLMs are evaluated by comparing their fine-tuned performance in downstream vision-language tasks. However, fine-tuning VLMs in downstream tasks only provides a black-box score, which renders poor interpretability of VLMs.

To acquire a deeper understanding of VLMs, a number of existing works have probed their capability from various perspectives, including verb understanding~\cite{DBLP:conf/acl/HendricksN21}, spatial relation understanding~\cite{DBLP:conf/naacl/RoschL22,DBLP:journals/corr/abs-2212-10015}, visual abstract reasoning with tangram shapes~\cite{DBLP:conf/emnlp/JiKRSVHA22}, generalization ability in out-of-domain datasets~\cite{DBLP:journals/corr/abs-2205-15237}, compositional reasoning ability~\cite{DBLP:conf/cvpr/ThrushJBSWKR22,DBLP:journals/corr/abs-2210-01936,DBLP:conf/nips/ParkALDR21}, visual-linguistic grounding capabilities on specific linguistic phenomena~\cite{DBLP:conf/acl/ParcalabescuCMF22}, robustness to image and text perturbations~\cite{DBLP:journals/corr/abs-2212-08044}, attribute recognition capability~\cite{DBLP:journals/corr/abs-2209-06103,DBLP:journals/corr/abs-2207-00221}, object hallucination problem~\cite{DBLP:journals/corr/abs-2210-07688}. 
These works have revealed that prevalent VLMs have severe shortcomings in certain aspects. 

Nevertheless, current probing works are still limited in the general domain. Specifically, they utilize images from the general domain to construct datasets or benchmarks, such as MSCOCO~\cite{DBLP:conf/eccv/LinMBHPRDZ14}, Visual Genome~\cite{DBLP:journals/ijcv/KrishnaZGJHKCKL17}, LAION-400M~\cite{DBLP:journals/corr/abs-2111-02114}, or social media data without domain specification. Instead of investigating VLMs in the general domain, we focus on VLMs' vision-language capability in the food domain, which is closely linked with people's health and daily life. To this end, we introduce a food image-caption dataset and comprehensively evaluate a range of representative VLMs on it

\subsection{Food Dataset}
\label{subsec:food_dataset}
In recent years, there have been substantial food datasets available. Most of them are proposed for food image classification, such as ETH Food-101 \cite{bossard14}, UPMC Food-101 \cite{DBLP:conf/icmcs/WangKTCP15} with western food, UEC Food256 \cite{DBLP:conf/eccv/KawanoY14}, Sushi-50 \cite{DBLP:conf/bmvc/QiuLSWL19} with Japanese food, VIREO Food-172 \cite{DBLP:conf/mm/ChenN16}, ChineseFoodNet \cite{DBLP:journals/corr/ChenZD17} with Chinese food, ISIA Food-500 \cite{DBLP:conf/mm/MinLWLWWJ20} comprising miscellaneous food categories worldwide. Besides category labels, UPMC Food-101 \cite{DBLP:conf/icmcs/WangKTCP15} and VIREO Food-172 \cite{DBLP:conf/mm/ChenN16} contain additional metadata such as related web text, ingredients, and cooking instructions. 

In addition, Yummly-66k \cite{DBLP:journals/tmm/MinBMZRJ18} annotates images with ingredients, courses and regions for food topic models. FoodSeg103 \cite{DBLP:conf/mm/WuFLLHS21} implements a food image segmentation dataset that tags each image with multiple ingredients and draws the corresponding pixel-wise masks. Recipe1M \cite{DBLP:journals/corr/abs-1810-06553} and Recipe1M+ \cite{DBLP:journals/pami/MarinBOHSAWT21} construct datasets with numerous images and recipes, which is suitable for image-recipe retrieval task. These datasets mainly facilitate specific tasks. 
However, the visual correlation between their textual annotations and images is relatively weak.
Their texts not only neglect plenty of visual attributes of the image but also contain invisible contents, e.g. cooking instructions~\cite{DBLP:journals/corr/abs-1810-06553,DBLP:journals/pami/MarinBOHSAWT21}. VLMs are hard to align these images and texts, which hinders these datasets from serving as probing datasets. To this end, we introduce Food-500 Cap, the first food image-caption dataset. Food-500 Cap has captions describing fine-grained visual content of the image. It also includes food category labels and their geographic origin tags. Hence, Food-500 Cap can serve as a comprehensive benchmark for probing VLMs' generalization ability in the food domain.

\begin{table*}[t]
\centering
\small
\resizebox{0.87\textwidth}{!}{
    \begin{tabular}{lccclc}
    \toprule
    \multirow{2}{*}{\textbf{Dataset}} & \multirow{2}{*}{\textbf{Image Number}} & \multicolumn{2}{c}{\textbf{Category}} & \multicolumn{2}{c}{\textbf{Annotation}} \\
    & & Number & Coverage & type & source \\
    \midrule
    Recipe1M+ \cite{DBLP:journals/pami/MarinBOHSAWT21} & 13M & - & - & Ingredients \& Cooking instructions & Web\\
    FoodSeg103 \cite{DBLP:conf/mm/WuFLLHS21} & 7,118 & 103 & Worldwide & Ingredients & Manual \\
    UPMC Food-101 \cite{DBLP:conf/icmcs/WangKTCP15} & 90,840 & 101 & Western & Related web text & Web\\
    UEC Food256 \cite{DBLP:conf/eccv/KawanoY14} & 25,088 & 256 & Japanese & - & - \\
    VIREO Food-172 \cite{DBLP:conf/mm/ChenN16} & 110,241 & 172 & Chinese & Ingredients \& Cooking instructions & Web \\
    Sushi-50 \cite{DBLP:conf/bmvc/QiuLSWL19} & 3,963 & 50 & Japanese & - & - \\
    ChineseFoodNet \cite{DBLP:journals/corr/ChenZD17} & 185,628 & 208 & Chinese & - & - \\
    Yummly-66k \cite{DBLP:journals/tmm/MinBMZRJ18} & 66,615 & - & - & Course \& ingredients \& region & Web \\
    ISIA Food-500 \cite{DBLP:conf/mm/MinLWLWWJ20} & 399,726 & 500 & Worldwide & - & - \\
    \midrule
    Food-500 Cap & 24,700 & 494 & Worldwide & Image Captions \& region & Manual \\
    \bottomrule
    \end{tabular}
}
\caption{Summary of popular food-domain datasets and Food-500 Cap. Our proposed Food-500 Cap has 24,700 images, covering 494 food categories around the world. Compared to existing food datasets, each image from Food-500 Cap has a hand-curated fine-grained image caption and the geographic origin of the food. Captions are annotated by a data annotation company and food origins are tagged by ourselves resorting to Wikipedia.}
\label{tb:dataset_comparison}
\end{table*}

\section{Food-500 Cap}
\label{sec:dataset}
This section outlines the construction process of the Food-500 Cap dataset and provides a detailed description of its statistics.

\subsection{Dataset Construction}
\paragraph{\textbf{Collecting food images.}} 
To ensure the diversity of food categories, we utilize images from the ISIA Food-500~\cite{DBLP:conf/mm/MinLWLWWJ20}, a comprehensive dataset for food recognition containing 399,726 samples covering 500 food categories from various countries and regions. 
For each category, we randomly select 50 images from ISIA Food-500. Note that actually we only use 494 out of the 500 categories currently and six categories are manually removed.

\paragraph{\textbf{Annotating food images.}} 
To obtain high-quality captions depicting fine-grained visual features, we employ a data annotation company and urge the annotators to follow the next three rules. 
First, annotators must include category labels in each caption, which contain the food's principal information. 
Second, we encourage annotators to be as careful as possible, marking all visible content of images including not only the food's color, shape, ingredients, seasonings, accessories, etc. but also the container's color, shape, pattern, etc. 
To ensure the distinctiveness of the captions, some general words should be avoided to the largest degree, such as \texttt{fruit} and \texttt{vegetables}. 
At last, every annotator is instructed to integrate the aforementioned information into fluent sentences using diverse syntactic constructions. 
Eventually, we obtain food image captions with fine-grained visual content and one example is shown in \Cref{fig:dataset_example}.

\paragraph{\textbf{Marking regions.}}
Although covering diverse food categories, one insufficiency of ISIA Food-500 is that it mixes food categories from different regions without marking their original regions, hindering further study of culinary cultures. Therefore, we resort to Wikipedia to mark the original region of each food category by ourselves, and we show the detailed process in \Cref{a:origins}. Consequently, all food categories are divided into seven regions: \textit{Worldwide}, \textit{Western}, \textit{Latin-American}, \textit{Chinese}, \textit{Japanese}, \textit{Indian}, and \textit{Asian}. \Cref{tb:dataset_regions} displays the food category distribution over these regions. Note that 90 food categories have ambiguous original places, so we merge them into \textit{Worldwide}.

\subsection{Dataset Statistics and Characteristics}
\label{subsec:dataset_statictics}
Food-500 Cap contains 24,700 images that are uniformly divided into 494 categories. Captions are of average length 18.57, and there are 7.26 nouns, 1.96 verbs, and 2.53 adjectives in each caption on average\footnote{https://spacy.io}. As shown in \Cref{tb:dataset_comparison}, Food-500 Cap surpasses current food datasets in the following two aspects: (1) Food-500 Cap annotates each image with a human-crafted, fine-grained, fluent visual description, hence containing richer visual information. (2) All food categories are tagged with their geographic origins, enabling culinary culture studies across regions. Whereas almost all current datasets neglect region annotation except for VIREO Food-172~\cite{DBLP:conf/mm/ChenN16}. Therefore, Food-500 Cap can better serve as a comprehensive vision-language benchmark. 

\section{Probing VLMs in Food Domain}
To comprehensively evaluate prevalent VLMs, we pick up three different types of VLMs including vision-language representation models~\cite{DBLP:conf/icml/RadfordKHRGASAM21, DBLP:conf/cvpr/SinghHGCGRK22, DBLP:conf/icml/0001LXH22, DBLP:conf/cvpr/YangDTXCCZCH22, DBLP:conf/icml/ZengZL22}, image-to-text generative models ~\cite{DBLP:conf/icml/0001LXH22,DBLP:conf/icml/WangYMLBLMZZY22,DBLP:journals/corr/abs-2205-14100}, and text-to-image generative models ~\cite{DBLP:conf/icml/RameshPGGVRCS21, DBLP:conf/cvpr/RombachBLEO22}. Then we apply four probing methods to them. For vision-language representation VLMs, we utilize classification and retrieval tasks to probe their cross-modal alignment ability. For generative VLMs, we utilize image captioning and image generation tasks to test whether they can generate satisfactory images or descriptions. All tasks are performed in a zero-shot setting to directly assess the generalization of VLMs.

\subsection{Vision-language Representation Models}
\subsubsection{Evaluated Models}
\label{subsubsec:evaluated_models}
\paragraph{\textbf{CLIP}
~\cite{DBLP:conf/icml/RadfordKHRGASAM21}} It employs two independent encoders to encode image and text respectively. It is trained with an image-text contrastive (ITC) objective, which encourages the embeddings of paired images and texts to be closer while pushing away those of mismatched pairs. Benefiting from 400 million image-text pairs during training, CLIP exhibits powerful zero-short transfer ability across a wide range of downstream tasks, e.g. cross-modal retrieval.
\paragraph{\textbf{TCL}~\cite{DBLP:conf/cvpr/YangDTXCCZCH22}} It consists of an image encoder, a text encoder, and a multi-modal encoder to fuse image and text features from uni-modal encoders. Apart from the original cross-modal contrastive objective like CLIP, TCL proposes intra-modal contrastive target and local mutual information maximization to robust the uni-modal representations. 
\paragraph{\textbf{X\_VLM}~\cite{DBLP:conf/icml/ZengZL22}} It shares the same model framework as TCL while better utilizing the region annotations in some datasets. It optimizes the model by predicting the location of bounding boxes in the image given the corresponding caption and meanwhile conducts vision-language alignment in multi-granularity.
\paragraph{\textbf{FLAVA}~\cite{DBLP:conf/cvpr/SinghHGCGRK22}} It inherits the architecture of TCL and X\_VLM. Different from VLMs only focus on cross-modal tasks, FLAVA is trained with regard to both cross-modal and uni-modal objectives, including global image-text contrastive learning, masked image modeling, masked language modeling, etc. And FLAVA achieves comparable results on vision-only, language-only, and cross-modal tasks.
\paragraph{\textbf{BLIP}~\cite{DBLP:conf/icml/0001LXH22}} It introduces a novel multi-modal mixture of Encoder-Decoder framework. It can operate as a uni-modal encoder, an image-grounded text encoder, and an image-grounded text decoder, sharing parameters with each other. They are optimized with contrastive loss, image-text matching (ITM) loss, and language modeling loss respectively. To leverage noisy web data effectively, BLIP augments the datasets utilizing captions synthesized by itself.

\begin{figure}[t]
	\centering
	\includegraphics[width=.49\textwidth] {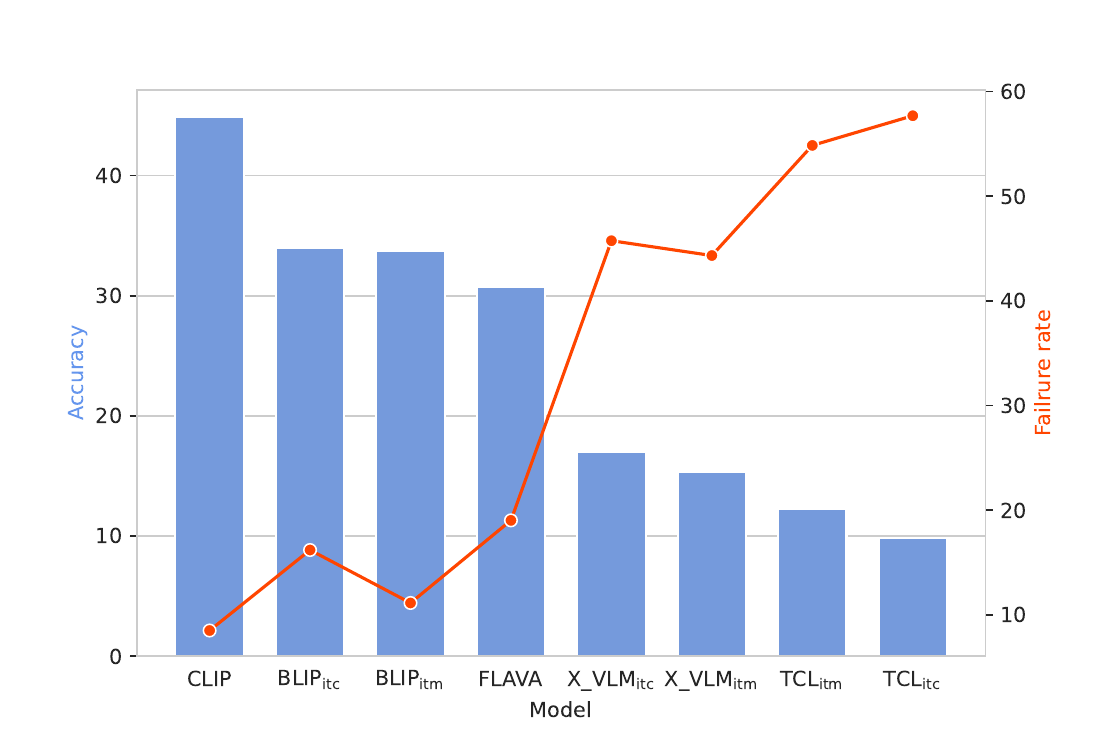}
	\caption{Results of zero-shot classification, including (1) accuracy  and (2) failure rate, which represents the percentage of categories where none of the images is correctly classified. Model names with subscript ``itm'' are of ITM configuration and others are of ITC configuration.}
\label{fig:zs_acc}
\end{figure}
\begin{table}[t]
\centering
\small
\resizebox{0.48\textwidth}{!}{
    \begin{tabular}{l|ccc|ccc}
    \toprule
    \multirow{2}{*}{\textbf{Model}} & \multicolumn{3}{c|}{\textbf{IR}} & \multicolumn{3}{c}{\textbf{TR}} \\
    
    & R@1 & R@5 & R@10 & R@1 & R@5 & R@10 \\
    \midrule
    CLIP & 7.51 & 20.77 & 30.32 & 9.28 & 23.98 & 33.63 \\
    FLAVA & 9.82 & 25.85 & 36.05 & 7.60 & 21.28 & 31.04 \\
    $\mathrm{{BLIP_{itc}}}$ & 15.28 & 33.12 & 42.37 & 15.90 & 34.15 & 44.02 \\
    $\mathrm{TCL_{itc}}$ & 3.32 & 9.25 & 13.55 & 2.19 & 6.65 & 10.23 \\
    $\mathrm{X\_VLM_{itc}}$ & 6.83 & 17.58 & 25.18 & 6.75 & 17.68 & 24.72 \\
    \midrule
    $\mathrm{BLIP_{itm}}$ & \textbf{24.43} & \textbf{46.01} & \textbf{55.22} & \textbf{22.09} & \textbf{43.21} & \textbf{53.57} \\
    $\mathrm{TCL_{itm}}$ & 8.63 & 18.29 & 23.91 & 7.11 & 14.98 & 19.40 \\
    $\mathrm{X\_VLM_{itm}}$ & 20.10 & 37.51 & 44.94 & 15.06 & 31.41 & 40.36 \\
    \bottomrule
    \end{tabular}
}
\caption{Zero-shot cross-modal retrieval results on Food-500 Cap. The overall best result is bold-face.}
\label{tb:whole_ret}
\end{table}

\begin{figure*}[t]
\centering
\subfigure[image classification accuracy]{\includegraphics[width=.49\textwidth]{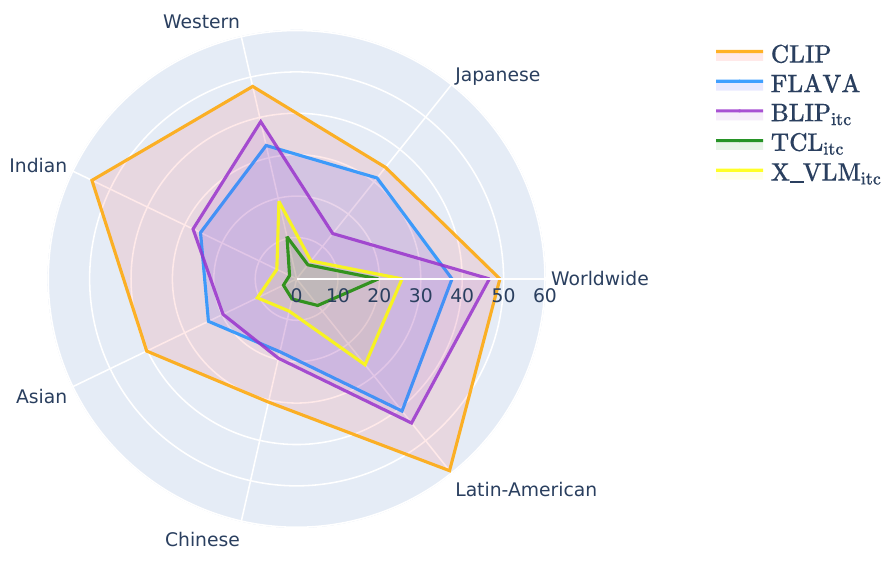}}
\subfigure[image-to-text retrieval R@1 score] {\includegraphics[width=.48\textwidth]{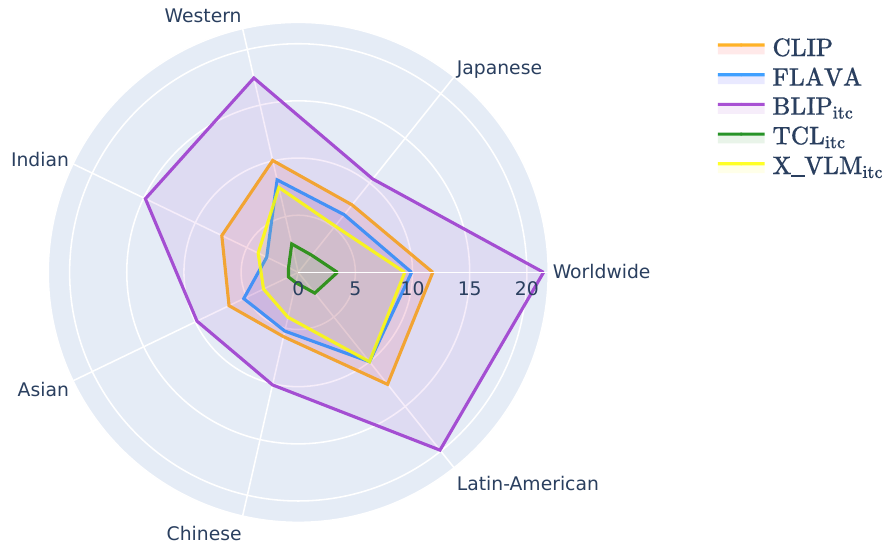}}
\caption{Radar chart of (a) Zero-shot image classification accuracy and (b) image-to-text retrieval R@1 score across regions. }
\label{fig:att_ret_ret_region}
\end{figure*}

\begin{figure}[t]
    \includegraphics[width=0.49\textwidth]{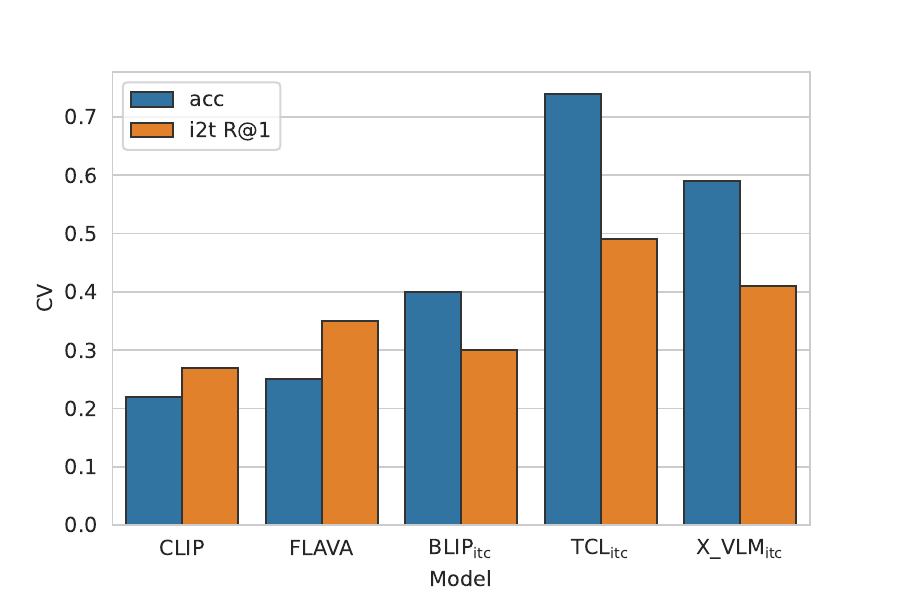}
    \caption{Coefficient of variation (CV) of classification accuracies (blue) and image-to-text retrieval R@1 scores (orange) across different regions of several VLMs. CV is the ratio of the standard deviation to the mean, which measures the dispersion of a probability distribution. In this figure, the higher the CV value, the more unbalanced performance across regions.}
    \label{fig:coeff_of_var}
\end{figure}

\subsubsection{Evaluation Task}
\paragraph{\textbf{Food Classification}}
Previous works~\cite{DBLP:conf/icml/RadfordKHRGASAM21,DBLP:conf/cvpr/SinghHGCGRK22} have revealed that VLMs have competitive zero-shot power in general-domain classification benchmarks, such as ImageNet~\cite{imagenet_cvpr09}, PASCAL VOC~\cite{DBLP:journals/ijcv/EveringhamGWWZ10}, CIFAR~\cite{krizhevsky2009learning}.
Furthermore, within the domain of food, CLIP~\cite{DBLP:conf/icml/RadfordKHRGASAM21} and FLAVA~\cite{DBLP:conf/cvpr/SinghHGCGRK22} report their overall accuracy on Food-101~\cite{bossard14}, but it has a relatively limited number of 101 food classes.
To this end, we employ representation VLMs to undertake zero-shot food classification using our benchmark and elaborate the results. 

Following ~\cite{DBLP:conf/icml/RadfordKHRGASAM21,DBLP:journals/ijcv/ZhouYLL22,DBLP:conf/emnlp/ShinRLWS20}, we undertake zero-shot food classification using prompt engineering. 
Specifically, we utilize a prompt template ``\texttt{a food photo of a \{label\}}'' and populate it with related category labels\footnote{We try several prompts and this one is of the overall highest accuracy.}. 
In the evaluation phase, we task the VLMs with identifying the correct prompt for each image among all constructed prompts. To be specific, we evaluate two configurations of vision-language representation models. The first is the ITC configuration, where only the image and text uni-modal encoders are employed. Images and prompts are individually embedded by VLMs' uni-modal encoders, and models select the prompt with the maximum cosine similarity for each image. The second is the ITM configuration, where BLIP, TCL, and X\_VLM further use the ITM score from their multi-modal fusion modules to re-rank the top-$k$ nearest prompts in the ITC configuration. Note that we fix $k$ to 128 and re-rank by adding the ITM score to cosine scores.

\paragraph{\textbf{Image-text Bidirectional Retrieval}}
Image-text bidirectional retrieval aims to retrieve images using textual queries (text-to-image retrieval) and converse (image-to-text retrieval). This task reveals models' ability to align the semantic space of vision and language. Though current VLMs~\cite{DBLP:conf/icml/RadfordKHRGASAM21, DBLP:conf/cvpr/YangDTXCCZCH22, DBLP:conf/icml/ZengZL22, DBLP:conf/cvpr/SinghHGCGRK22, DBLP:conf/icml/0001LXH22} have achieved superior zero-shot performance on general domain datasets such as Flickr~\cite{DBLP:journals/tacl/YoungLHH14}, MSCOCO~\cite{DBLP:conf/eccv/LinMBHPRDZ14}, it is unknown whether they also perform well in a specific domain. 
Consequently, we conduct zero-shot image-text bidirectional retrieval using the food image-caption pair in Food-500 Cap. Like food classification. Similar to food classification, this task is also performed using ITC and ITM configurations. Finally, we report the top-$1$, $5$, and $10$ retrieval scores on our benchmark.

\subsubsection{Results}

\begin{table*}[t]
\centering
\small
\resizebox{0.99\textwidth}{!}{
\begin{tabular}{l|l|c|ccccccc}
\toprule
\textbf{Setting} & \textbf{Model} & \textbf{Overall} & \textbf{\textit{Chinese}} & \textbf{\textit{Worldwide}} & \textbf{\textit{Japanese}} & \textbf{\textit{Western}} & \textbf{\textit{Latin-American}} & \textbf{\textit{Asian}} & \textbf{\textit{Indian}} \\
\midrule
 \multirow{3}{*}{Accurate} & $\mathrm{BLIP_{Dec}}$ & 5.29 & 1.37 & 8.38 & 0 & 7.43 & 8.11 & 0.29 & 0 \\
 & $\mathrm{GIT}$ & 2.98 & 1.33 & 4.49 & 0 & 3.98 & 6.00 & 0.20 & 0 \\
 & $\mathrm{OFA}$ & \textbf{6.85} & \textbf{4.87} & \textbf{9.99} & \textbf{0.82} & \textbf{7.98} & \textbf{11.47} & \textbf{3.26} & \textbf{0.24} \\
\midrule
 \multirow{3}{*}{Relaxed} & $\mathrm{BLIP_{Dec}}$ & \textbf{30.76} & 29.83 & \textbf{45.19} & 0.45 & \textbf{37.32} & 20.53 & 11.55 & 4.12 \\
 & $\mathrm{GIT}$ & 26.27 & \textbf{31.70} & 36.43 & 0.73 & 30.56 & 16.74 & 11.06 & 5.30 \\
 & $\mathrm{OFA}$ & 29.51 & 30.70 & 39.32 & \textbf{3.82} & 35.74 & \textbf{20.74} & \textbf{12.58} & \textbf{7.07} \\
\bottomrule
\end{tabular}
}
\caption{Semantic Label Accuracy (\%) in the entire dataset (Overall) and different regions according to our taxonomy of food categories. \textit{Accurate}: the generated caption exactly contains the whole food category label. \textit{Relaxed}: the generated caption contains some word from the food label.}
\label{tb-caption_region_acc}
\end{table*}
\begin{table}[t]
\centering
\small
\resizebox{0.49\textwidth}{!}{
\begin{tabular}{l|l|ccccc}
\toprule
\textbf{Model} & Avg. Len & B@4 & M & R & C & CLIP-S \\
\midrule
$\mathrm{BLIP_{Dec}}$ & 13.80 & 2.61 & 8.71 & \textbf{20.33} & 13.62 & 0.70 \\
$\mathrm{GIT}$ & 20.89 & 2.00 & 8.81 & 16.78 & 9.92 & 0.70 \\
$\mathrm{OFA}$ & 20.45 & \textbf{2.64} & \textbf{9.14} & 17.89 & \textbf{14.01} & 0.71 \\
\midrule
GT & 18.57 & - & - & - & - & \textbf{0.78} \\
\bottomrule
\end{tabular}
}
\caption{Results of image captioning on various metric. B@4, M, R C and CLIP-S represent BLEU@4, METEOR, ROUGE, CIDEr and CLIPScore respectively.}
\label{tb:caption_result}
\end{table}

\paragraph{\textbf{Classification}} 
\Cref{fig:zs_acc} displays the overall food classification results. None of the models achieve an accuracy above 50\%. CLIP achieves over 40\% accuracy on the zero-shot food classification task, which has the highest accuracy. BLIP and FLAVA also show competitive performance, while TCL and X\_VLM exhibit a sizable gap. This phenomenon should be attributed to their relatively small pre-training datasets, which only contain 4M\footnote{TCL does not release its checkpoint pre-trained on 16M image-text pairs.} and 16M image-text pairs, while that for FLAVA, BLIP, and CLIP are 70M, 129M, and 400M, respectively.
We also obtain the following findings:

\textbf{VLMs fail to recognize certain food categories.} 
The accuracy varies greatly in different food categories. 
On one hand, VLMs perform nearly perfectly in some categories. 
For example, CLIP correctly classifies all images in \textit{bandeja paisa}  from \textit{Latin-American}. However, On the other hand, VLMs recognize no images from some categories, such as \textit{aburaage} and \textit{doufunao}. 
The percentage of such categories for all VLMs is displayed in \Cref{fig:zs_acc} using the failure rate metric. 
We notice that even the best-performing CLIP fails in nearly 10\% of categories. And both X\_VLM and TCL fail to identify over 40\% categories.

\textbf{VLMs exhibit culinary culture bias in zero-shot food classification.}
As illustrated in \Cref{fig:att_ret_ret_region} (a), all models exhibit consistency in their performance across different regions. 
These models can better identify food images from \textit{Western} and \textit{Latin-American} than others except for \textit{Worldwide}.
Besides the qualitative results, we furthermore report the coefficient of variation (CV) of scores in different regions in \Cref{fig:coeff_of_var}, which reveals strong culinary culture bias in TCL and $\mathrm{X\_VLM}$.
Such bias is probably inherited from the pre-training dataset, where food from some countries or regions, e.g. Japan, appears much less frequently than European and American food.

\paragraph{\textbf{Image-text Bidirectional Retrieval}}
\Cref{tb:whole_ret} shows the overall performance of the compared models on image-text bidirectional retrieval. 
Similar to our findings in food classification, the results of retrieval also demonstrate VLMs underperform in the food domain compared to the general domain and suffer from the region bias. To be specific:

\textbf{The overall performance is not satisfactory.}
For the ITC configuration, BLIP gets the highest score on R@1, but it only reaches 15\%. Other models get scores below 10\%. 
In contrast to the classification task, where CLIP achieved the highest accuracy, it does not perform well in retrieval tasks. This implies that while CLIP is better at recognizing the general type of food, it has a weaker ability to distinguish food at a fine-grained level.
Using the ITM configuration, this problem can be alleviated a bit. Under this setting, BLIP, TCL, and X\_VLM obtain much higher scores on R@1, R@5, and R@10.

\textbf{All VLMs also suffer in certain categories in image-text retrieval.}
For example, $\mathrm{BLIP_{itm}}$'s image-to-text recall@1 score reaches 66.0 for \textit{christmas cake}, a food from \textit{Western}, but it hardly retrieves correct descriptions for \textit{bon bon chicken} which is from \textit{Chinese}. We further investigate the retrieval results in different regions.
As shown in \Cref{fig:att_ret_ret_region} (b), we find that all VLMs perform relatively poorly in \textit{Asian}, \textit{Chinese}, \textit{Indian} and \textit{Japanese}.  According to the quantitative results in \Cref{fig:coeff_of_var}, CLIP also maintains the lowest CV, which suggests a relatively weak culinary culture bias. We speculate the reason to be its tremendous amount of pre-training data.

\subsection{Image-to-text Generative Models}
\subsubsection{Evaluated Models}
\paragraph{\textbf{GIT}~\cite{DBLP:journals/corr/abs-2205-14100}} It is composed of one swin-like~\cite{DBLP:conf/iccv/LiuL00W0LG21} vision transformer and one text decoder. During training, it uses the language modeling task to predict the associated caption given an image. When applied to downstream tasks, GIT first transforms them into text generation and then produces the answer word by word. 
\paragraph{\textbf{OFA}~\cite{DBLP:conf/icml/WangYMLBLMZZY22}} It proposes a more generic encoder-decoder framework compared with GIT. It develops a unified multi-modal vocabulary, and both its encoder and decoder can process inputs from different modalities. Therefore, OFA can serve as a task-agnostic and modality-agnostic model. Both multi-modal and uni-modal tasks are combined during pre-training, which renders OFA superior performance on a wide range of tasks, such as image captioning, image generation, image classification, language understanding, etc.
\paragraph{\textbf{BLIP}~\cite{DBLP:conf/icml/0001LXH22}} Owing to its special architecture, BLIP can be regarded as an image-grounded text decoder as well. Hence we use BLIP in this part and denote it as $\mathrm{BLIP_{Dec}}$. Detailed introduction can be referred to \Cref{subsubsec:evaluated_models}.

\begin{figure*}[t]
\centering
\includegraphics[width=.99\textwidth] {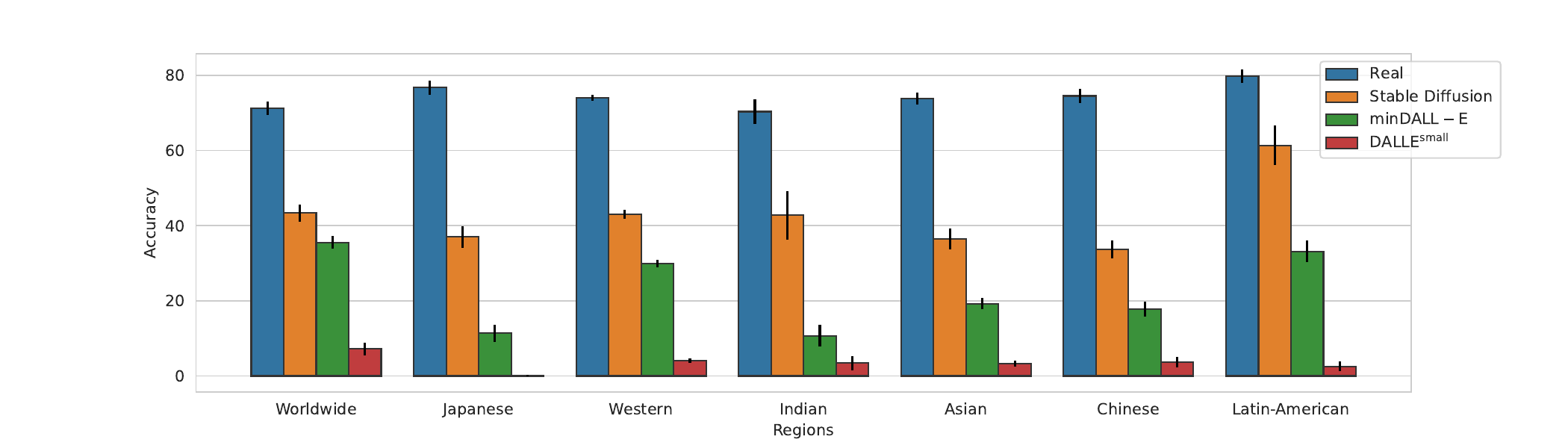}
\caption{Per-region accuracy of the classifier trained on real vs. synthetic images. We randomly split the dataset 10 times and reported the test accuracy on real images.}
\label{fig:linear_acc_regions}
\end{figure*}

\subsubsection{Evaluation Task}
\paragraph{\textbf{Image Captioning}}
Image-to-text generative VLMs aggregate multiple tasks into a unified text generation task~\cite{DBLP:conf/icml/0001LXH22, DBLP:journals/corr/abs-2205-14100, DBLP:conf/icml/WangYMLBLMZZY22}.
Thus we opt to leverage zero-shot image captioning as a means of probing these models. 
The objective of image captioning is to generate descriptive sentences for given images.
We utilize multi-view metrics to better exhibit the generative ability of VLMs. First, we calculate common-used image captioning metrics, including n-gram-based metrics, such as BLEU~\cite{lin2004rouge}, METEOR~\cite{DBLP:conf/acl/PapineniRWZ02}, ROUGE~\cite{DBLP:conf/wmt/DenkowskiL14} and CIDEr~\cite{DBLP:conf/cvpr/VedantamZP15}, and a semantic-based metric, CLIPScore~\cite{DBLP:conf/emnlp/HesselHFBC21}.

Then, we assess the recognition ability of VLMs. Inspired by Semantic Object Accuracy (SOA)~\cite{DBLP:journals/pami/HinzHW22} in text-to-image generation evaluation, which evaluates whether the generated image contains the objects mentioned in the text, we check whether the generated captions contain the food labels of the images, and similarly define Semantic Label Accuracy (SLA): 
\begin{equation*}
\mathrm{SLA}=\frac{1}{N}\sum_{i=1}^N\mathbf{1}(l^{(i)}\in c^{(i)})   
\end{equation*}
where $N$ is the number of images, $l^{(i)}$ and $c^{(i)}$ are the category label and generated caption of the $i$th image respectively.

\subsubsection{Results}
\ 
\newline \indent We provide the captioning results in \Cref{tb:caption_result} including the n-gram-based metrics score and CLIPScore for all three models. All models have low scores on n-gram-based metrics.

\textbf{All image-to-text generative models hardly generate the fine-grained attributes}. 
From a perspective of metrics, the low scores on n-gram-based metrics indicate a clear literal mismatch between the generated and reference captions that contain many fine-grained attributes, including the shape, color, and ingredient. 
We provide some generated descriptions in \Cref{tb:caption_example}. From the Table, we can observe that all VLMs neglect or misidentify a lot of food attributes. 
In the top example, models do not generate \textit{eggplants}, \textit{potatoes}, \textit{green peppers} compared to the reference in our proposed dataset. Moreover, we find that OFA tends to generate hallucinations~\cite{DBLP:conf/emnlp/RohrbachHBDS18, DBLP:journals/corr/abs-2210-07688}, and GIT prefers to append meaningless words to the end of the sentence. 
As shown in the bottom example of \Cref{tb:caption_example}, OFA generates ``\textit{ready to be served to the guests at the wedding reception}'', and GIT adds ``\textit{yum}'' and invalid punctuation at the end of the caption. These phenomena lead to the longer captions of GIT and OFA compared with $\mathrm{BLIP_{Dec}}$ (\Cref{tb:caption_result}), but there is no significant advantage in caption metrics. 
They also have lower CLIP-S scores than the ground truth, suggesting the weaker alignment of the generated captions with the images. 

\textbf{These models hardly generate correct labels in descriptions}
The low scores on metrics may reflect the generated sentence including incorrect labels. 
To verify this, we  display the overall SLA and that in different regions in \Cref{tb:caption_region_acc}, revealing that less than 10\% captions exactly include whole category labels for all three models. 
If we relax the requirement and regard it as true if the generated caption contains some word in the label (\textit{Relaxed} setting in \Cref{tb:caption_region_acc}) rather than the whole label, SLA obtains significant improvement for all models, especially in \textit{Chinese}, \textit{Worldwide} and \textit{Western}.
This is because many category labels (e.g. \textit{lentil soup}) from these regions contain common words like \texttt{soup}, which is easier for VLMs to generate. 

\textbf{Food labels from different regions pose different levels of difficulty for VLMs to generate.}
As shown in \Cref{tb:caption_region_acc}, VLMs can hardly generate food category labels from specific regions, indicating a possible bias in culinary culture. 
In particular, for \textit{Japanese} and \textit{Indian}, $\mathrm{BLIP_{Dec}}$ and GIT fail to generate food labels from these countries, and even the highest performing OFA only achieves 0.82 and 0.24 SLA, respectively. 

\begin{table}[t]
\centering
\small
\resizebox{0.49\textwidth}{!}{
    \begin{tabular}{cccc}
    \toprule
    \textbf{Model} & \textbf{FID ($\downarrow$)} & $\bm{\mathrm{FID_{CLIP}}}$ ($\downarrow$) & \textbf{CAS ($\uparrow$)} \\ 
    \midrule
    Stable Diffusion & 25.74 & 10.92 & 41.45 ± 0.89 \\
    minDALL-E & 28.64 & 15.46 & 26.60 ± 0.92\\
    $\mathrm{DALLE^{small}}$ & 54.49 & 29.91 & 4.21 ± 0.54 \\
    \midrule
    Real & 0 & 0 & 73.79 ± 0.69 \\
    \bottomrule
    \end{tabular}
}
\caption{FID, $\bm{\mathrm{FID_{CLIP}}}$ and CAS for different text-to-image generation models. For CAS, we randomly split the test data 10 times and report the mean and standard deviation.}
\label{tb:draw_quality}
\end{table}

\subsection{Text-to-Image Generative Models}
\subsubsection{Evaluated Models}
\paragraph{\textbf{DALL-E}~\cite{DBLP:conf/icml/RameshPGGVRCS21}} It employs a decoder-only transformer that receives texts and images as a single stream. Given a text prompt, it first predicts the image tokens autoregressively, which have been pre-defined in the codebook of a pre-trained discrete variational autoencoder (dVAE)~\cite{DBLP:conf/iclr/Rolfe17}. Then the generated image tokens are fed to the decoder of the dVAE to synthesize images. Training on 250M image-text pairs from the internet, DALL-E can create plausible images for various sentences even in the zero-shot setting. Note that the checkpoint is unavailable. We choose two publicly released implements: minDALL-E~\cite{kakaobrain2021minDALL-E}, $\mathrm{DALL\-E^{small}}$\footnote{https://github.com/lucidrains/DALLE-pytorch}.

\paragraph{\textbf{Stable Diffusion}~\cite{DBLP:conf/cvpr/RombachBLEO22}} It leverages the prevalent diffusion model, which learns to reverse the process of adding noise to images. Unlike diffusion-based AI painters such as GLIDE~\cite{DBLP:conf/icml/NicholDRSMMSC22}, Imagen~\cite{DBLP:journals/corr/abs-2205-11487}, Stable Diffusion uses latent diffusion model. The latent diffusion model operates in a compressed image space rather than the high-dimensional pixel space. Consequently, Stable Diffusion can generate high-resolution images from text descriptions with less computation consumption.

\subsubsection{Evaluation Task}
\paragraph{\textbf{Image Synthesis}} Following the default implementation, we adopt Stable Diffusion, minDALL-E, and $\mathrm{DALLE^{small}}$ to synthesize images given food descriptions. 
To evaluate the overall image quality, we use Fr\'echet Inception Distance (FID)~\cite{DBLP:conf/nips/HeuselRUNH17} and $\mathrm{FID_{CLIP}}$~\cite{kynkaanniemi2022role} scores, which measures the distance between the distributions of the real and synthetic images in the feature space of an ImageNet pre-trained Inception-v3~\cite{DBLP:conf/cvpr/SzegedyVISW16} and CLIP~\cite{DBLP:conf/icml/RadfordKHRGASAM21}, respectively. Then we employ Classification Accuracy Score (CAS)~\cite{DBLP:conf/nips/RavuriV19} to assess to what extent the generated images manifest the categorical condition, where generated images are used to train a classifier which is then used to predict the label of real images. To compute CAS, a classifier is first trained on the generated images, then used to predict labels of real images. 

\subsubsection{Results}
\ 
\newline \indent
We report the FID and $\mathrm{FID_{CLIP}}$ score and the CAS score in \Cref{tb:draw_quality}, which shows that all three models exhibit significant differences in their performance on image synthesis tasks. 
Through quantitative and qualitative analysis, we find the following issues:

\textbf{There is a significant gap between synthesis and real images.}
From \Cref{tb:draw_quality}, all models have a significant gap compared to real images on FID, $\mathrm{FID_{CLIP}}$, which measure the similarity between the synthesis and real images. Especially, the performance of $\mathrm{DALLE^{small}}$ is worse than the other two models. \Cref{fig:draw_samples} shows some generated synthetic images. Through \Cref{fig:draw_samples}, we find that images generated by $\mathrm{DALLE^{small}}$ are unrealistic. In contrast, those generated by Stable Diffusion appear relatively more realistic and contain more caption content. 
To investigate whether the models can capture the main features of the category mentioned in the text, we further provide a quantitative evaluation of the synthetic images. 
Unlike metrics such as FID and $\mathrm{FID_{CLIP}}$, CAS ignores some fringe features and is concerned with whether the generated images contain the necessary features to represent the class. We find that all models suffer a performance drop compared to real images, which indicates that text-to-image generative models might have difficulty capturing representative category features. 

\textbf{Text-to-image Generative models also suffer from region imbalance issues, similar to the previous models.}
For all regions, we observe a sizeable accuracy gap between the synthetic and real images. For example, Stable Diffusion's accuracy drops from nearly 20 to 40 across all regions. Furthermore, as shown in \Cref{fig:linear_acc_regions}, all models have some culinary culture bias. Specifically, the classifier trained with images generated by Stable Diffusion achieves particularly higher accuracy in \textit{Latin-American} than other regions. And minDALL-E scores higher in \textit{Worldwide}, \textit{Western}, \textit{Latin-American} than Asian countries. As for $\mathrm{DALLE^{small}}$, it fails to recognize almost all food images from \textit{Japanese}. In contrast, using real images to train the classifier results in relatively balanced accuracy across regions, which might be because those text-to-image generative models are trained on a biased dataset, which contains fewer traditional food types from certain regions such as \textit{Chinese} and \textit{Japanese}.

\section{Conclusion}
In this work, we introduce Food-500 Cap, a new vision-language benchmark in the food domain.
By in-house labeling, Food-500 Cap not only provides each image with a fine-grained visual content description but also labels a novel taxonomy that divides food categories into their geographic origins, which aids in studying different culinary cultures.
We adopt four vision-language tasks in the zero-shot setting, including food classification, image-text bidirectional retrieval, image captioning, and image synthesis, and evaluate nine VLMs of three different architectures on our proposed benchmark. 
Experiments reveal VLMs' limitations in the food domain and their bias against culinary culture.
We hope that our proposed benchmark will promote the study of multi-modal food computing and our findings will provide insights into the deployment and application of VLMs in the food domain.

\begin{acks}
We would like to thank the anonymous reviewers for their insightful comments. Shujian Huang and Jianbing Zhang are the corresponding authors. This work is supported by National Science Foundation of China (No. 62176120 and No. 62176115), the Liaoning Provincial Research Foundation for Basic Research (No. 2022-KF-26-02).
\end{acks}

\clearpage
\bibliographystyle{ACM-Reference-Format}
\balance
\bibliography{software}

\clearpage
\appendix
\onecolumn
\section{Process of annotating geographic origins}
\label{a:origins}
\begin{figure}[H]
\centering
\subfigure[Wonton noodles]{\includegraphics[width=.22\textwidth]{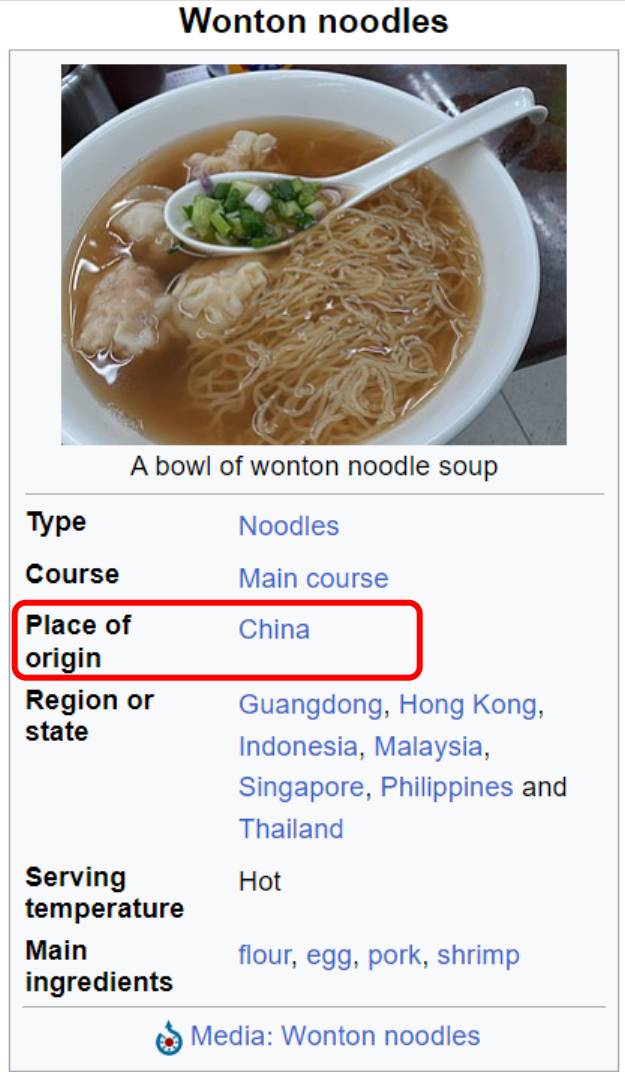}}\hspace{8mm}
\subfigure[Tsukemono] {\includegraphics[width=.72\textwidth]{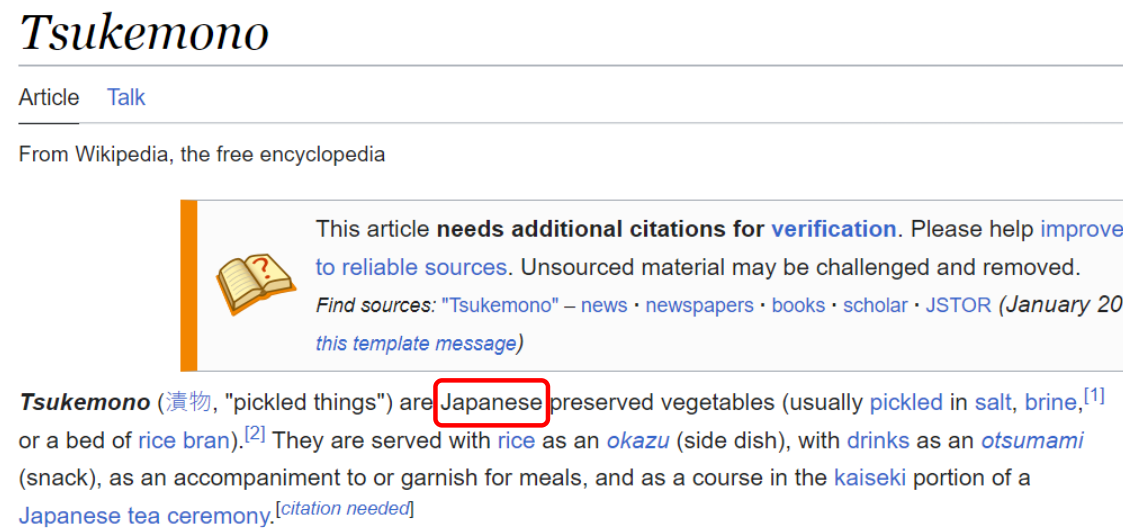}}
\caption{Annotating geographic origins in the case that the clear origin of the food are given. (a) \textit{Wonton noodles}, where the place of origin is directly provided. (b) \textit{Tsukemono}, where the entry lacks additional citations for verification and the place of origin is not provided, but we can still find the origin in the article.}
\label{fig:annotate_1}
\end{figure}

\begin{figure}[H]
\centering
\includegraphics[width=.90\textwidth]{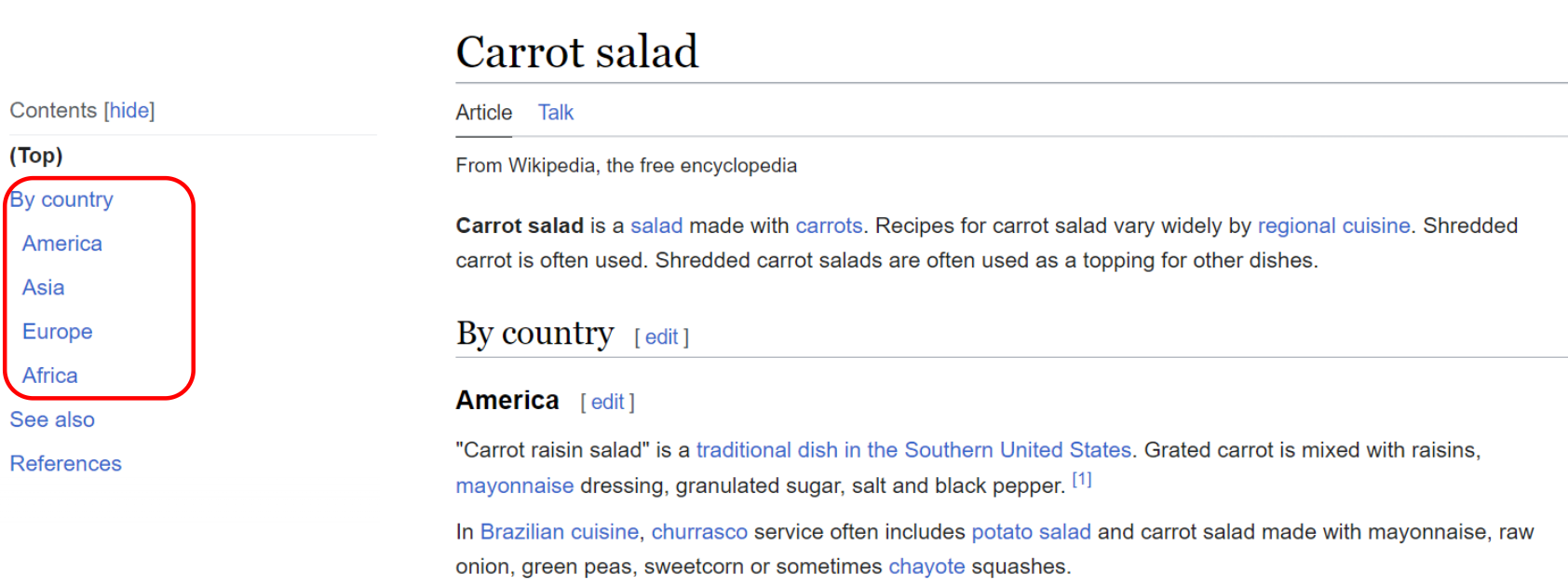}
\caption{Annotating geographic origins when some food categories have unknown places of origin.}
\label{fig:annotate_2}
\end{figure}

For each food category, we resort to its Wikipedia entry. We can find their places of origin of most food categories. We display two examples in \Cref{fig:annotate_1}, according to which we assign \textit{Wonton noodles} to region \textit{Chinese} and \textit{Tsukemono} to region \textit{Japanese}. However, some food categories have unknown origins, such as the \textit{Carrot salad} shown in \Cref{fig:annotate_2}. We assign these food categories to \textit{Worldwide} in our taxonomy.

\begin{table}[htbp]
\centering
\small
\resizebox{0.50\textwidth}{!}{
    \begin{tabular}{lcl}
    \toprule
    \textbf{Region} & \textbf{\#Categories} & \textbf{Covered Countries} \\
    \midrule
    \textit{Worldwide} & 90 & unknown original regions \\
    \textit{Western} & 216 & Europe \& America \& Canada \\
    \textit{Latin-American} & 19 & Latin American countries \\
    \textit{Chinese} & 60 & China \\
    \textit{Japanese} & 22 & Japan \\
    \textit{Indian} & 17 & India \\
    \textit{Asian} & 70 & Asia except for China, Japan, India \\
    \bottomrule
    \end{tabular}
}
\caption{Region distribution of food categories in Food-500 Cap.}
\label{tb:dataset_regions}
\end{table}

\begin{table}[h!]
\centering
\resizebox{0.49\textwidth}{!}{
\scriptsize
\begin{tabular}{c|l}
\toprule
\textbf{Image} & \textbf{Captions} \\
\midrule
\multirow{8}{*}{\begin{minipage}{.105\textwidth}\includegraphics[width=\linewidth]{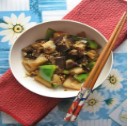}\end{minipage}}
& \underline{\textbf{GT}} di san xian sauteed with soft eggplants, potatoes \\
& and slices of green peppers in a bowl which placed\\
& on a red napkin.\\
\cmidrule{2-2}
& \underline{$\bm{\mathrm{BLIP_{Dec}}}$} a bowl of food with chops and chops on \\
& a blue and white floral tablecloth. \\ 
& \underline{\textbf{OFA}} a bowl of food with chopsticks on a blue and\\ 
& white tablecloth with white daisies in the background. \\ 
& \underline{\textbf{GIT}} chicken in a bowl with chopsticks on a red and \\
& blue placemat. yum!!! yum yum... \\
\midrule
\multirow{8}{*}{\begin{minipage}{.11\textwidth}\includegraphics[width=\linewidth]{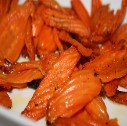}\end{minipage}}
& \underline{\textbf{GT}} a grilled piece of bone-in pork knuckle served with \\
& yellow sauerkraut, and decorated with rosemary. \\
\cmidrule{2-2}
& \underline{$\bm{\mathrm{BLIP_{Dec}}}$} a plate of carrots on a table.\\ 
& \underline{\textbf{OFA}} a plate of sweet potato fries with a drizzle of \\
& olive oil on top, ready to be served to the guests at \\
& the wedding reception.\\
& \underline{\textbf{GIT}} cooked carrots in a white plate with a brown \\
& sauce... yum!!! : - ) ( : - - - )\\
\bottomrule
\end{tabular}
}
\caption{Examples of generated captions from three image-to-text generative models compared to ground truth (GT).}
\label{tb:caption_example}
\end{table}

\begin{figure}[h!]
\centering
\includegraphics[width=.49\textwidth] {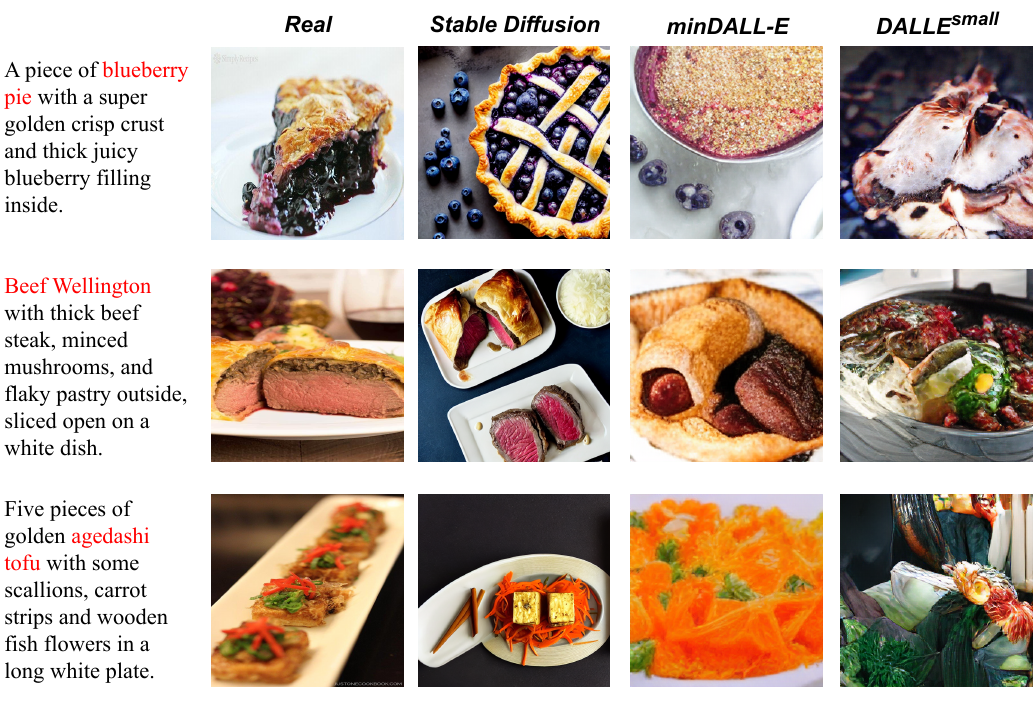}
\caption{Some examples of real images and synthesis images from text-to-image generative models}
\label{fig:draw_samples}
\end{figure}

\end{document}